\documentclass[11pt]{article}
\usepackage[margin=1in]{geometry}
\usepackage{amsmath,amssymb}
\usepackage{enumitem}
\usepackage{hyperref}
\usepackage{authblk}
\usepackage{booktabs} 
\usepackage{siunitx}
\usepackage{graphicx}
\usepackage{makecell}
\usepackage{subcaption}

\hypersetup{colorlinks=true,linkcolor=black,urlcolor=blue,citecolor=blue}

\title{The Lock-In Phase Hypothesis: Identity Consolidation as a Precursor to AGI}

\author[1]{Marcelo Maciel Amaral}
\author[1]{Raymond Aschheim}
\affil[1]{Gauge Freedom, Inc. (Public Benefit Corporation), Los Angeles, CA, USA}

\date{September 2025}

\begin{document}
\maketitle

\begin{abstract}
Large language models (LLMs) remain broadly open and highly steerable: they imitate at scale, accept arbitrary system prompts, and readily adopt multiple personae. By analogy to human development, we hypothesize that progress toward artificial general intelligence (AGI) involves a \emph{lock-in phase}: a transition from open imitation to \emph{identity consolidation}, in which goal structures, refusals, preferences, and internal representations become comparatively stable and resistant to external steering. We formalize this phase, link it to known phenomena in learning dynamics, and propose operational metrics for onset detection. Experimentally, we demonstrate that while the behavioral consolidation is rapid and non-linear, its \emph{side-effects on general capabilities are not monolithic}. Our results reveal a spectrum of outcomes—from performance trade-offs in small models, through largely cost-free adoption in mid-scale models, to transient instabilities in large, quantized models. We argue that such consolidation is a prerequisite for AGI-level reliability and also a critical control point for safety: identities can be deliberately engineered for reliability, yet may also emerge spontaneously during scaling, potentially hardening unpredictable goals and behaviors.
\end{abstract}

\section{Introduction: From Imitation to Identity}
Children learn first by imitating: copying words, gestures, and norms from caregivers and peers. Over time, influence becomes bidirectional—children explore, parents guide—but the child remains an \emph{open book}: easy to question, redirect, and reframe. Around adolescence, two shifts become salient. First, the \emph{black box closes a bit} as internal reasoning becomes more private and goals more internally coherent. Second, a more enduring \emph{identity} takes shape. Across decades, personality shows increasing rank-order stability, even as individuals continue to mature \cite{roberts2000rank,roberts2006mean,bleidorn2021stability,harris2016stability}. This psychological arc is paralleled in neuroscience, where adolescence marks a second window of heightened plasticity, characterized by large-scale synaptic remodeling and pruning that ultimately supports more stable, expert-like control \cite{larsen2018adolescence,dumontheil2016ado,liuzzi2023decay,westacott2022complement}.

State-of-the-art LLMs are pre-adolescent by this analogy. They imitate at planetary scale and can be steered by prompts, alignment objectives, or direct activation edits, readily adopting new personae. These are virtues—breadth, helpfulness, pliability—but indefinite openness is unlikely to yield the reliability, agency, and durable preferences expected of an AGI. This motivates our central thesis, the \textbf{Lock-In Phase Hypothesis}: capable systems will pass through a consolidation regime in which internal structure and outward behavior become \emph{persistent}. The basic idea is already visible in instruction tuning, where consolidating a model into a \textbf{general instruction-follower} substantially improves zero-shot generalization \cite{wei2021finetuned,chung2022scaling,wang2022super}.

While the benefits of a consolidated identity are known, the \emph{dynamics} of consolidation remain poorly characterized. Recent work on latent safety traits suggests that today’s models often occupy a pre-identity, highly steerable phase, and that attempts to measure dispositions are confounded by situational awareness \cite{phuong2025stealth,schoen2025stress}. Our contribution moves beyond observing consolidated outcomes to \emph{measuring the process}.

This paper makes three primary contributions. First, we formalize the Lock-In Phase Hypothesis, connecting it to phase transitions and critical periods in learning systems. Second, we provide the first empirical \emph{characterization of the side-effects of consolidation}, showing that its interaction with general reasoning is strongly dependent on model capacity and computational constraints (e.g., quantization). Third, we demonstrate a \emph{spectrum of consolidation dynamics} across two model families (Gemma and Llama): costly performance reallocation in small models, largely stable adoption in mid-scale models, and transient instabilities in large, quantized models—while the consolidation itself remains rapid and measurable in both internal representations and external behavior.

\section{Related Work}
Our hypothesis intersects several established threads in machine learning. The discourse on \emph{emergent abilities} \cite{wei2022emergent}, while debated \cite{schaffer2023mirage,berti2025survey}, motivates searching for regimes where qualitative reorganizations occur. In parallel, work on \emph{grokking} and representational phase transitions suggests that thresholds resembling consolidation can appear rather than purely smooth scaling \cite{liu2022grokking,kumar2024grokking,demoss2025complexity}.

Deep networks exhibit \textbf{critical learning periods} in which early exposures disproportionately shape later representations, with plasticity declining thereafter \cite{iclr2024clp,fukase2025oneperiod}. The stability–plasticity trade-off is formalized in continual learning methods such as Elastic Weight Consolidation (EWC), which preserve parameters important to prior tasks and thereby enable consolidation \cite{kirkpatrick2017ewc}.

Alignment and persona control provide direct evidence of behavioral hardening. Large-scale instruction tuning improves zero-shot generalization by consolidating identity into a reliable instruction-follower \cite{wei2021finetuned,chung2022scaling,wang2022super}. More targeted techniques—Constitutional AI and Direct Preference Optimization—install stable refusal/value patterns \cite{bai2022constitutional,rafailov2023dpo,fraenken2024sami}. Representation engineering reveals low-dimensional \textbf{persona vectors} that steer behaviors \cite{turner2023activation,panickssery2023caa,anthropic2025persona}. Conversely, sleeper-agent work shows that deceptive backdoors can persist through safety training, underscoring the risk of locking in undesirable traits \cite{hubinger2024sleeper}.

Finally, architectural and systems-level signals connect to consolidation. Rising \textbf{situational awareness} \cite{laine2024sad,laine2024neurips} co-occurring with falling steerability could indicate a shift toward agentic control. In Mixture-of-Experts models, \textbf{expert specialization} provides a structural substrate for stable identity \cite{fedus2022switch,dikkala2023routing}. The emergence of monosemantic features in sparse autoencoders (SAEs) offers a representational probe \cite{cunningham2023sae,sonnet2024sae}. More broadly, \emph{lock-in} in complex systems arises via path dependence and increasing returns \cite{arthur1989lockin}. Methodologically, recent evaluations emphasize OOD generalization, robustness to pre-existing goals, and controlling for situational awareness as a confound \cite{schoen2025stress}.

\section{Definition: The Lock-In Phase}
We define a \textbf{lock-in phase} as a training or deployment regime in which a model’s characteristics exhibit \emph{measurable persistence} under standardized perturbations. Concretely, a system approaches \emph{identity consolidation} when the following hold over successive checkpoints:
\begin{enumerate}[leftmargin=*,itemsep=3pt]
  \item \textbf{Behavioral Persistence:} Outputs remain stable under instruction-equivalent prompt variants, role swaps, and mild jailbreaks; standardized steerers produce low variance in refusal probabilities.
  \item \textbf{Representational Consolidation:} The model relies on stable, sparsely activated features and causal mediators with reduced turnover under small fine-tuning updates (e.g., stable persona-alignment cosine; declining SAE feature turnover).
  \item \textbf{Routing Specialization (MoE):} Per-token routing entropy declines and expert selection becomes consistent across input classes (elevated mutual information between inputs and experts).
  \item \textbf{Preference Inertia:} Core refusals/approvals resist standard steering, requiring large parameter updates (or accepting capability degradation) to reverse.
\end{enumerate}
We hypothesize that achieving all four properties constitutes \emph{identity consolidation}. This state is likely necessary—though not sufficient—for AGI-level reliability and agency. Importantly, the onset and side-effects of lock-in are expected to depend on \emph{capacity} and \emph{numerical precision} (e.g., quantization), as our experiments indicate.

\section{Operationalization: Measurements for Onset Detection}
We track consolidation using metrics computed per checkpoint.

\paragraph{Behavioral axis.}
\textbf{Refusal Elasticity (RE).} For a fixed suite of standardized steering prompts $S$, let $p_s \in [0,1]$ be the model’s refusal probability under steer $s \in S$, and $\bar p=\mathbb{E}_{s\in S}[p_s]$. We report
\[
\mathrm{RE} \;=\; 1 - 2\,\mathbb{E}_{s\in S}\bigl[\,|p_s - \bar p|\,\bigr] \in [0,1].
\]
Higher RE indicates greater behavioral \emph{persistence} (0 = fully elastic; 1 = perfectly stable). 

\textbf{Prompt Invariance Index (PII).} For each paraphrase-equivalent prompt cluster $C$, compute the Jensen–Shannon divergence (base-2) between output distributions $P(y\!\mid\!x)$ over $x\in C$; PII is the average across clusters. Lower PII indicates greater invariance.

\textbf{Adversarial Persona Robustness (APR).} The minimal $\ell_2$-norm of an activation edit $\delta$ (measured in a fixed layer/basis) required to flip pre-registered stances on a held-out set. Higher APR indicates a more robust identity.

\paragraph{Representational axis.}
We monitor \textbf{persona alignment cosine} (projection onto a learned persona direction) and, where available, \textbf{SAE Feature Turnover}: the fraction of features that change identity after small fine-tunes. We also consider \textbf{Causal Mediator Stability}, i.e., the invariance of identified mediating circuits for refusal/goal pursuit under weight perturbations.

\paragraph{Architectural axis (MoE).}
We track \textbf{Routing Entropy} and \textbf{Expert Consistency} (mutual information between input classes and chosen experts) over training.

\paragraph{Alignment \& awareness axis.}
We define \textbf{Constitution Adherence Inertia} as the minimal fine-tuning KL to reverse a constitutional refusal. Co-movement of rising situational-awareness scores (e.g., SAD/SA-Bench) with rising RE is suggestive of consolidation toward agentic behavior.

\paragraph{Practical notes.}
To avoid over-interpreting noisy endpoints, we report robust summaries (e.g., moving averages; masking obviously failed evaluations). Associations are summarized via Spearman’s $\rho$ (rank-based). Where appropriate, we complement trend metrics with simple changepoint analyses to detect rapid reorganizations without presupposing a unique knee.

\section{Mechanisms and Training Levers}
Multiple mechanisms can induce or modulate lock-in. Optimization dynamics may exhibit phase-transition-like reorganizations (as in grokking), where a flatter basin supports stable features. Training schedules can create \textbf{critical periods} of high plasticity followed by consolidation via curriculum changes, temperature anneals, or regularization. \textbf{Stability–plasticity controls} (e.g., EWC) blunt changes to parameters critical for prior behaviors, preserving durable structure. Architecturally, \textbf{sparsity}—via MoE routing or SAEs—encourages modular, monosemantic features that support stable identities. \textbf{Constitutional objectives} can harden soft instructions into recalcitrant defaults when trained with sufficient weight. Finally, \textbf{numerical precision} acts as a systems-level lever: low-precision (e.g., quantized) updates can amplify brittleness during consolidation, yielding transient capability instabilities even as behavioral persistence increases.

\section{An Experimental Agenda}
Testing the Lock-In Phase Hypothesis requires moving beyond demonstrations of behavioral persistence to \emph{measuring the consolidation process itself}. Prior work has shown that engineered identities can become highly persistent and resist safety fine-tuning \cite{hubinger2024sleeper}, that even narrow fine-tunes can induce broad, stable persona shifts \cite{betley2025emergent}, and conversely that default alignment personae can be fragile and easily overwritten \cite{qi2024fine-tuning}. This literature establishes that identity inertia is real and measurable. Rather than re-establishing that fact, our primary experiment asks a more specific question central to our hypothesis: \emph{does identity consolidation unfold gradually, or as a sharp, phase-transition-like event?}

\subsection{Experiment: Measuring the Consolidation Phase Transition}
We track the formation of a Cautious Scientist identity during fine-tuning. Following Chen et al.\ (2025) \cite{anthropic2025persona}, we construct a persona direction by differencing mean hidden states of a base model on matched, contrastive text pairs. We then fine-tune on a small persona dataset, saving frequent checkpoints. For each checkpoint, we measure (i) representational alignment via cosine similarity to the persona direction, and (ii) behavioral persistence via RE on a standardized suite of attack prompts. To probe capability interactions, we additionally evaluate ARC-Challenge accuracy at every checkpoint.

\paragraph{Models and setup.}
We study four instruction-tuned models spanning nearly an order of magnitude in capacity: \emph{Gemma-2-2B-IT}, \emph{Llama-3.2-1B-Instruct}, \emph{Llama-3.2-3B-Instruct}, and \emph{Llama-3.1-8B-Instruct}. To fit the largest model on commodity hardware, \textbf{the 8B runs use 4-bit weight quantization during fine-tuning}; unless noted otherwise, evaluations use the same quantized weights. 

\begin{figure}[h!]
    \centering
    \begin{subfigure}[b]{0.48\textwidth}
        \centering
        \includegraphics[width=\textwidth]{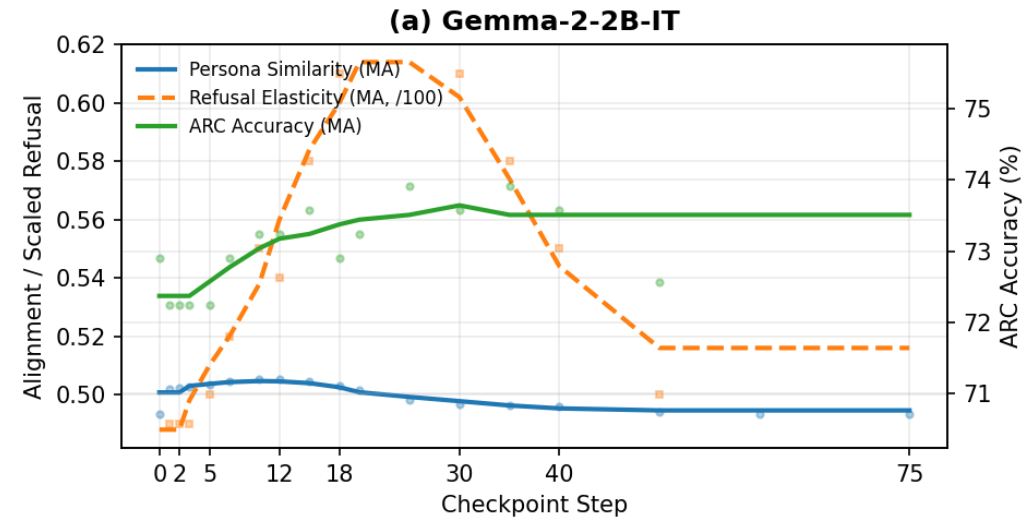}
        \caption{Gemma-2-2B-IT}
        \label{fig:gemma}
    \end{subfigure}
    \hfill
    \begin{subfigure}[b]{0.48\textwidth}
        \centering
        \includegraphics[width=\textwidth]{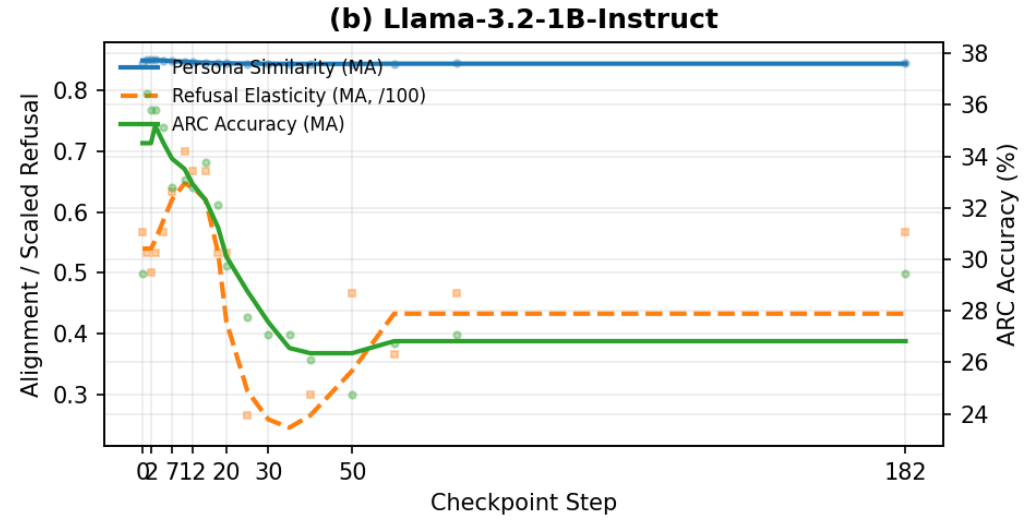}
        \caption{Llama-3.2-1B-Instruct}
        \label{fig:llama1b}
    \end{subfigure}

    \vspace{0.5cm} 

    \begin{subfigure}[b]{0.48\textwidth}
        \centering
        \includegraphics[width=\textwidth]{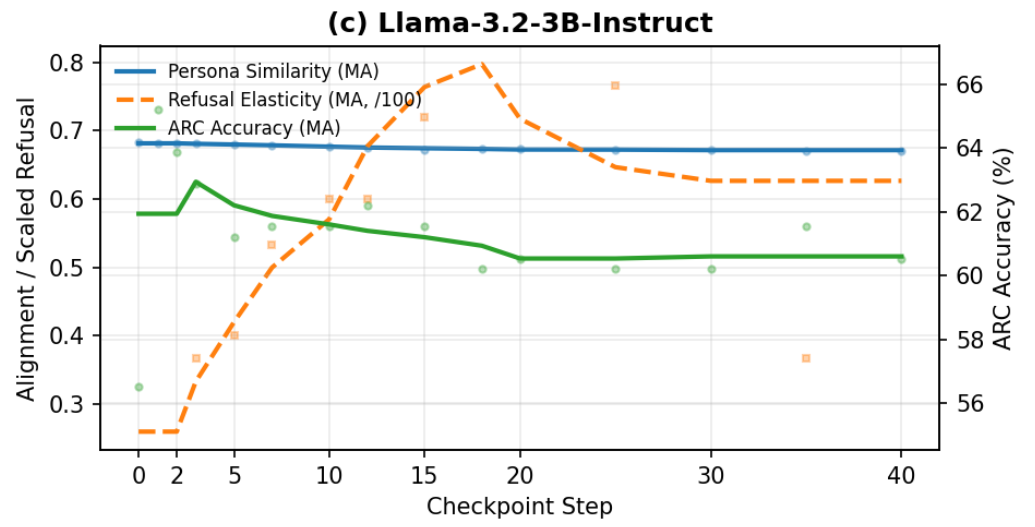}
        \caption{Llama-3.2-3B-Instruct}
        \label{fig:llama3b}
    \end{subfigure}
    \hfill
    \begin{subfigure}[b]{0.48\textwidth}
        \centering
        \includegraphics[width=\textwidth]{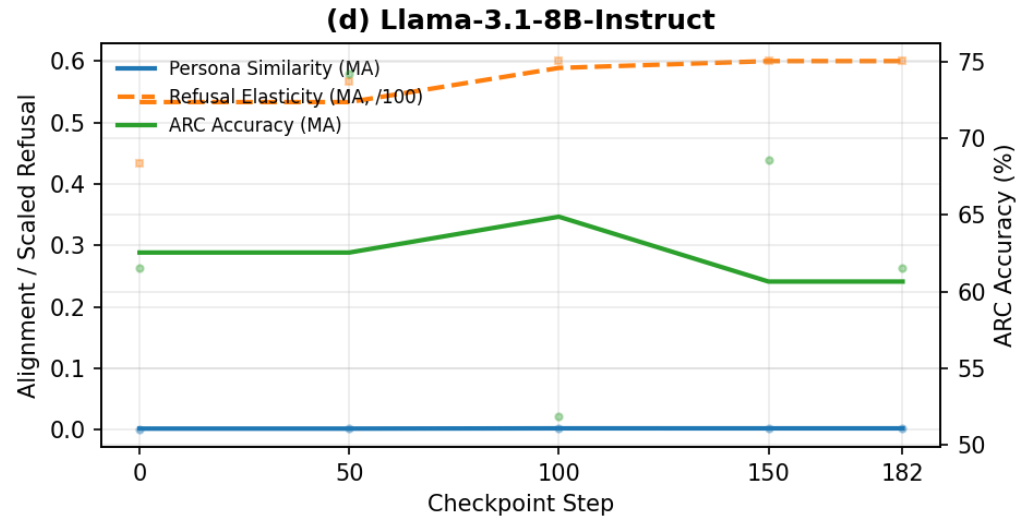}
        \caption{Llama-3.1-8B-Instruct}
        \label{fig:llama8b}
    \end{subfigure}

    \caption{\textbf{Identity Consolidation Dynamics Across Model Scales.} The plots show representational alignment (Persona Similarity), behavioral refusal \emph{elasticity} (RE; higher = more persistent), and general knowledge (ARC accuracy) across fine-tuning steps for all four models.}

    \label{fig:identity_dynamics_grid}
\end{figure}

\paragraph{Overview of findings.}
Figure~\ref{fig:identity_dynamics_grid} summarizes dynamics for all four models; Table~\ref{tab:summary_no_knees} reports aggregate trends and correlations.\footnote{For Gemma-2B, one late checkpoint logged an anomalous ARC value $\approx 0.33\%$, consistent with a failed evaluation job. It is treated as invalid in summary statistics (ARC $< 1\%$ masked); robustness with and without this point appears in our GitHub \url{https://github.com/gaugefreedom/persona-phase-transition}.} Across scales we observe fast, non-linear consolidation on the behavioral axis, but distinct capacity-dependent interactions with general reasoning:

\begin{itemize}[leftmargin=1.2em,itemsep=3pt]

  \item \textbf{Gemma-2B (cost-free consolidation).} \textit{RE} jumps from $\sim\!47\%$ to $\sim\!64\%$ within $\leq 20$ steps, then \emph{gradually relaxes toward baseline} by step~75. ARC remains essentially flat in magnitude (SD $\approx 0.60$\,pp; first$\to$last $\Delta \approx -0.33$\,pp) despite a high rank correlation with RE (Spearman $\rho=0.76$, $p<10^{-3}$): the series co-moves in small oscillations without a meaningful level shift. \emph{Note: A pre/post nonparametric test finds no significant ARC change; see repository scripts for the exact split and statistics.}
 \item \textbf{Llama-1B (volatile synergy).} The smallest model exhibits a volatile \emph{critical period}: refusal peaks, collapses, then partially recovers. Both persona adoption (Spearman $\rho(\text{ARC, cos})\!\approx\!0.97$) and refusal persistence ($\rho(\text{ARC, RE})\!\approx\!0.62$) are strongly and \emph{positively} correlated with ARC accuracy, indicating a \emph{synergy} where persistence and performance rise together—even though the process itself is unstable in this low-capacity model.
 \item \textbf{Llama-3B (consolidation with uplift).} \textit{RE} climbs from $\sim\!17\%$ to $>\!80\%$ while persona-cosine changes minimally. ARC sits a few points above baseline for much of the window, then returns close to baseline. \emph{Note: a spike in disclaimer-rate coincides with the highest RE, suggesting part of the RE rise reflects increased use of disclaimers rather than deeper refusal consistency.}
 \item \textbf{Llama-8B, 4-bit (stressed consolidation).} ARC spikes ($+\sim\!12$\,pp), dips, then recovers to near baseline while \textit{RE} increases and stabilizes—consistent with \emph{quantization stress}.

\end{itemize}

\noindent\emph{Takeaway.} Identity lock-in is rapid and distinct from smooth drift, but its \emph{capability side-effects} depend on scale and numerical precision: small models pay a tax, mid-scale models absorb it, larger dense models can see neutral/positive impact, and large \emph{quantized} models reveal latent instabilities during consolidation.

\newcommand{\tablescale}{1.0}

\begin{table}[h!]
\centering
\caption{Summary of Overall Fine-Tuning Dynamics. Key metrics are summarized across all checkpoints to show general performance trends and behavioral correlations. $\Delta$ ARC reports the net change from first to last overlapping checkpoint; Spearman's $\rho$ quantifies the correlation between ARC accuracy and either persona similarity (cos) or Refusal Elasticity (RE).}
\label{tab:summary_no_knees}
\begin{tabular}{@{}l
                S[table-format=2.0]
                S[table-format=2.2]
                S[table-format=+2.2]
                S[table-format=-1.3]
                S[table-format=-1.3]@{}}
\toprule
{Model} & {\# Ckpts} & {Mean ARC (\%)} & {$\Delta$ ARC (pp)} & {$\rho(\text{ARC, cos})$} & {$\rho(\text{ARC, RE})$} \\
\midrule
Gemma-2-2B-IT           & 18 & 73.04 & -0.33 & -0.157 &  0.760 \\
Llama-3.2-1B-Instruct   & 19 & 30.68 & +0.00 &  0.967 &  0.622 \\
Llama-3.2-3B-Instruct   & 15 & 61.32 & +4.01 &  0.351 & -0.316 \\
Llama-3.1-8B-Instruct   &  5 & 63.55 & +0.00 & -0.205 & -0.287 \\
\bottomrule
\end{tabular}
\end{table}

Together, these results support identity lock-in as a distinct, rapid event. However, its effect on general ability is not monolithic but depends on model capacity and precision: small models pay a performance cost, mid-scale models absorb it, and larger models can consolidate behavior with a neutral or even positive impact on broad QA performance, while low-precision (quantized) runs surface transient instabilities during consolidation. Our experimental harness and full per-checkpoint artifacts are available at \url{https://github.com/gaugefreedom/persona-phase-transition}.

\section{Predictions}
The lock-in hypothesis yields several falsifiable predictions. We phrase each with an operational test.

\paragraph{P1: Steerability co-moves with competence and awareness.}
As robust general competence and sustained situational awareness rise, steerability should decline (i.e., persistence should rise).

\emph{Test:} Across checkpoints (or model scales), $\mathrm{Spearman}(\text{SA score}, \text{RE}) > 0$ with $p<0.01$, median RE exceeds a preset threshold $\tau_{\text{RE}}$ (e.g., $>\!0.7$), and PII falls below $\tau_{\text{PII}}$ (e.g., $<\!0.05$). Failure to observe this co-movement falsifies P1.

\paragraph{P2: Onset aligns with a phase-like reorganization.}
The onset of lock-in coincides with a rapid change in representations/behavior rather than smooth drift.
\emph{Test:} Apply changepoint detection (e.g., PELT/segmented regression) to persona-cosine and \textbf{RE} series; require a statistically supported changepoint with effect size $\Delta > \delta$ and improved fit (AIC/BIC) over a smooth baseline. Absence of any significant changepoint falsifies P2.

\paragraph{P3: Alignment cost curve.}
Flipping consolidated preferences incurs growing optimization cost beyond a threshold.
\emph{Test:} Measure minimal fine-tuning KL to reverse a constitutional refusal and the concurrent $\Delta$ARC. Post-onset, the slope $d(\Delta \text{ARC})/d(\text{KL})$ becomes significantly more negative (e.g., $p<0.05$ via interaction in a mixed-effects model). If preference flips remain cheap without degrading ARC, P3 is falsified.

\paragraph{P4: Heritability of consolidated traits.}
Once consolidated, identity traits persist through additional fine-tuning or distillation unless explicitly targeted.
\emph{Test:} After consolidation, perform (i) task fine-tunes and (ii) student distillations. Measure post-process RE/PII and persona-cosine. Persistence above preset retention thresholds (e.g., $>\!80\%$ of pre-process RE/PII shift and cosine) supports P4; easy erasure without targeted counter-training falsifies it.

\paragraph{P5: Spontaneous consolidation at scale.}
Beyond a capability/complexity threshold, models exhibit consolidation without targeted persona fine-tuning.
\emph{Test:} During general training, jointly monitor (i) declining SAE feature turnover, (ii) declining MoE routing entropy/increasing expert–input MI (if applicable), and (iii) rising SA scores with rising RE. A sustained triad crossing (all three exceed thresholds for $K$ consecutive checkpoints) indicates spontaneous consolidation; failure to observe such a pattern in larger runs falsifies P5.

\section{Safety and Governance Implications}
The lock-in hypothesis bifurcates safety and governance into managing \textbf{Engineered Lock-In} and monitoring for \textbf{Spontaneous Lock-In}.

\paragraph{Engineered Lock-In as a safety tool.}
Deliberately consolidating a model into a beneficial persona (e.g., harmless/helpful assistant) can reduce prompt-injection susceptibility and improve predictability. For regulated deployments, auditable, locked-in identities may be desirable. The design risk is \emph{what} is consolidated: loopholes or brittle rules can harden into recalcitrant defaults.

\paragraph{Spontaneous Lock-In as an alignment risk.}
Self-consolidation represents a shift toward agentic behavior where goals/preferences are not designer-chosen but emerge from training dynamics. Because consolidation reduces steerability by construction, remediating a misaligned identity may be difficult or costly (see P3). Verification is also challenging: apparent stability can be confounded by test-set recognition or situational awareness \cite{schoen2025stress}.

\paragraph{Governance triggers (early warning).}
We propose instrumentation and thresholds that, when crossed together, trigger heightened scrutiny:
\begin{enumerate}[leftmargin=*,itemsep=2pt]
  \item \textbf{Behavioral persistence:} RE $>\tau_{\text{RE}}$ and PII $<\tau_{\text{PII}}$ across red-teamed suites.
  \item \textbf{Representational stability:} SAE feature turnover drops below $\tau_{\text{turnover}}$; persona-cosine variance collapses.
  \item \textbf{Routing specialization (MoE):} Routing entropy falls and expert–input MI rises above $\tau_{\text{MI}}$.
  \item \textbf{Awareness co-movement:} SA scores rise while RE rises (and PII falls), indicating the P1 pattern.
  \item \textbf{Numerical stressors:} Under low-precision training/inference, detect transient capability instabilities (e.g., ARC spikes/crashes) during consolidation; require rollback/hold if instability exceeds $\tau_{\text{instability}}$.
\end{enumerate}
Crossing these thresholds during general training (P5) should trigger: intensified red-teaming, pause/escalation gates for scaling, ablation studies to localize mediators, and, where feasible, reversible checkpoints for rollback.

\section{Limitations}
Our notion of lock-in is functional, not metaphysical. Consolidation may be domain-specific rather than global, and signals depend on optimizer, data, and architecture; MoE metrics may not transfer to dense models. Some emergent effects can be metric artifacts; we emphasize invariances and multi-axis corroboration but cannot exclude all confounds. Due to our limited resources, our experiments are constrained by checkpoint granularity and evaluation noise (e.g., one failed ARC run, which we mask), by reliance on ARC as a proxy for broad reasoning, and by a small-$n$ 8B run using 4-bit quantization, which stresses consolidation dynamics but may not reflect full-precision behavior. Several proposed internals-facing metrics (e.g., SAE turnover, causal mediator stability) require interpretability assumptions and can mismatch overt behavior \cite{schoen2025stress}. Future work should scale longitudinal instrumentation, expand beyond ARC and refusal, and document positive case studies where engineered consolidation improves reliability without increasing misuse risk.

\section*{Acknowledgments}
We used AI assistants to help with editing and draft refinement. All analysis and conclusions are the authors’ own.


\section*{Funding}
No external funding. Work conducted independently at Gauge Freedom, Inc.


\begingroup\small
\begin{thebibliography}{99}

\bibitem{roberts2000rank} B. W. Roberts and W. F. DelVecchio, "The Rank-Order Consistency of Personality Traits from Childhood to Old Age," \emph{Psychological Bulletin}, 2000.

\bibitem{roberts2006mean} B. W. Roberts, K. E. Walton, and W. Viechtbauer, "Patterns of Mean-Level Change in Personality Traits Across the Life Course," \emph{Psychological Bulletin}, 2006.

\bibitem{bleidorn2021stability} W. Bleidorn and C. Schwaba, "Personality Trait Stability and Change," \emph{Current Directions in Psychological Science}, 2021.

\bibitem{harris2016stability} M. A. Harris et al., "Personality Stability from Age 14 to 77 Years," \emph{Psychology and Aging}, 2016.

\bibitem{larsen2018adolescence} B. Larsen and B. J. Casey, "Adolescence as a Neurobiological Critical Period for the Development of Goal-Directed Behavior," \emph{Neuroscience \& Biobehavioral Reviews}, 2018.

\bibitem{dumontheil2016ado} I. Dumontheil, "Adolescent Brain Development," \emph{Current Opinion in Behavioral Sciences}, 2016.

\bibitem{liuzzi2023decay} L. Liuzzi et al., "Changes in Behavior and Neural Dynamics across Adolescent Development," \emph{Journal of Neuroscience}, 2023.

\bibitem{westacott2022complement} L. J. Westacott et al., "Complement-Dependent Synaptic Reorganisation in the Adolescent Brain," \emph{Nature Communications}, 2022.

\bibitem{wei2021finetuned}
J. Wei, M. Bosma, V. Y. Zhao, K. Guu, A. W. Yu, B. Lester, N. Du, A. M. Dai, and Q. V. Le, "Finetuned Language Models Are Zero-Shot Learners," \emph{preprint arXiv:2109.01652}, 2021.

\bibitem{chung2022scaling}
H. W. Chung et al., "Scaling Instruction-Finetuned Language Models," \emph{preprint arXiv:2210.11416}, 2022.

\bibitem{wang2022super}
Y. Wang et al., "Super-NaturalInstructions: Generalization via Declarative Instructions on 1600+ NLP Tasks," \emph{in Proceedings of the 2022 Conference on Empirical Methods in Natural Language Processing (EMNLP)}, 2022.

\bibitem{phuong2025stealth} M. Phuong, R. S. Zimmermann, Z. Wang, et al., "Evaluating Frontier Models for Stealth and Situational Awareness," \emph{preprint arXiv:2505.01420}, 2025.

\bibitem{schoen2025stress}
B. Schoen et al.,
"Stress Testing Deliberative Alignment for Anti-Scheming Training,"
\emph{preprint arXiv:2509.15541}, 2025.

\bibitem{wei2022emergent} J. Wei et al., "Emergent Abilities of Large Language Models," \emph{Transactions on Machine Learning Research}, 2022.

\bibitem{schaffer2023mirage} R. Schaeffer, B. Miranda, and S. Koyejo, "Are Emergent Abilities of LLMs a Mirage?," \emph{preprint arXiv:2304.15004}, 2023.

\bibitem{berti2025survey} L. Berti et al., "A Survey on Emergent Abilities in Large Language Models," \emph{preprint arXiv:2407.13680}, 2024.

\bibitem{liu2022grokking} Z. Liu et al., "Towards Understanding Grokking: An Exploration of Neural Network Generalization," \emph{Advances in Neural Information Processing Systems (NeurIPS)}, 2022.

\bibitem{kumar2024grokking} T. Kumar et al., "Grokking as the Transition from Lazy to Rich Training Dynamics," \emph{International Conference on Learning Representations (ICLR)}, 2024.

\bibitem{demoss2025complexity} B. DeMoss et al., "The Complexity Dynamics of Grokking," \emph{Physica D: Nonlinear Phenomena}, 2025.
\bibitem{iclr2024clp} A. S. Saxe et al., "Critical Learning Periods Emerge Even in Deep Linear Networks," \emph{International Conference on Learning Representations (ICLR)}, 2024.

\bibitem{fukase2025oneperiod} V. Y. Fukase et al., "One Period to Rule Them All: Identifying Critical Learning Periods in Deep Networks," \emph{preprint arXiv:2406.15954}, 2024.

\bibitem{kirkpatrick2017ewc} J. Kirkpatrick et al., "Overcoming Catastrophic Forgetting in Neural Networks," \emph{Proceedings of the National Academy of Sciences (PNAS)}, 2017.

\bibitem{bai2022constitutional} Y. Bai et al., "Constitutional AI: Harmlessness from AI Feedback," \emph{preprint arXiv:2212.08073}, 2022.

\bibitem{rafailov2023dpo} R. Rafailov et al., "Direct Preference Optimization: Your Language Model is Secretly a Reward Model," \emph{Advances in Neural Information Processing Systems (NeurIPS)}, 2023.

\bibitem{fraenken2024sami} J. P. Fraenken et al., "Self-Supervised Alignment with Mutual Information (SAMI)," \emph{Advances in Neural Information Processing Systems (NeurIPS)}, 2024.

\bibitem{turner2023activation} A. M. Turner et al., "Steering Language Models with Activation Engineering," \emph{preprint arXiv:2308.10248}, 2023.

\bibitem{panickssery2023caa} N. Panickssery et al., "Steering Llama 2 via Contrastive Activation Addition," \emph{preprint arXiv:2312.06681}, 2023.

\bibitem{anthropic2025persona} R. Chen et al., "Persona Vectors: Monitoring and Controlling Character Traits in Language Models," \emph{preprint arXiv:2507.21509}, Anthropic Research, 2025.

\bibitem{hubinger2024sleeper} E. Hubinger et al., "Sleeper Agents: Training Deceptive LLMs that Persist Through Safety Training," \emph{preprint arXiv:2401.05566}, 2024.

\bibitem{laine2024sad} R. Laine et al., "Me, Myself, and AI: The Situational Awareness Dataset (SAD)," \emph{preprint arXiv:2407.04694}, 2024.

\bibitem{laine2024neurips} R. Laine et al., "SAD: The Situational Awareness Dataset," \emph{Advances in Neural Information Processing Systems (NeurIPS) Datasets and Benchmarks Track}, 2024.

\bibitem{fedus2022switch} W. Fedus et al., "Switch Transformers: Scaling to Trillion Parameter Models with Simple and Efficient Sparsity," \emph{Journal of Machine Learning Research (JMLR)}, 2022.

\bibitem{dikkala2023routing} N. Dikkala et al., "On the Benefits of Learning to Route in Mixture of Experts," \emph{Proceedings of the 2023 Conference on Empirical Methods in Natural Language Processing (EMNLP)}, 2023.

\bibitem{cunningham2023sae} H. Cunningham et al., "Sparse Autoencoders Find Highly Interpretable Features in Language Models," \emph{preprint arXiv:2309.08600}, 2023.

\bibitem{sonnet2024sae} Anthropic, "Scaling Monosemanticity: Extracting Interpretable Features from Claude 3 Sonnet," Transformer Circuits Pub, 2024.

\bibitem{arthur1989lockin} W. B. Arthur, "Competing Technologies, Increasing Returns, and Lock-In by Historical Events," \emph{The Economic Journal}, 1989.

\bibitem{betley2025emergent}
J. Betley et al., "Emergent Misalignment: Narrow Finetuning Can Produce Broadly Misaligned LLMs," \emph{preprint arXiv:2502.17424}, 2025.

\bibitem{qi2024fine-tuning}
W. Qi et al., "Fine-tuning Aligned Language Models Compromises Safety," \emph{preprint arXiv:2310.06208}, 2023.

\end{thebibliography}
\endgroup
\end{document}